\documentclass[letterpaper]{article}
\usepackage{AA-SIPP-ArXiv}
\usepackage{graphicx}
\usepackage{amssymb}
\usepackage{amsmath}
\usepackage{multirow}
\usepackage{float}
\usepackage{times}
\usepackage{helvet}
\usepackage{courier}
\usepackage{color}
\usepackage{textcomp}
\usepackage[linesnumbered, ruled, vlined]{algorithm2e}
\SetKwProg{Fn}{Function}{}{}

\title{Any-Angle Pathfinding for Multiple Agents Based on SIPP Algorithm}
\author{Konstantin Yakovlev\\ 
Federal Research Center \\ ``Computer Science and Control''\\
of Russian Academy of Sciences, \\Moscow, Russia \\
National Research University Higher School of Economics\\
Moscow, Russia\\
yakovlev@isa.ru \And Anton Andreychuk\\
RUDN University,\\ Moscow, Russia \\
Federal Research Center \\ ``Computer Science and Control''\\
of Russian Academy of Sciences, \\Moscow, Russia \\
andreychuk@mail.com }

\begin{document}

\maketitle
\begin{abstract}

The problem of finding conflict-free trajectories for multiple agents of identical circular shape, operating in shared 2D workspace, is addressed in the paper and decoupled, e.g., prioritized, approach is used to solve this problem. Agents' workspace is tessellated into the square grid on which any-angle moves are allowed, e.g. each agent can move into an arbitrary direction as long as this move follows the straight line segment whose endpoints are tied to the distinct grid elements. A novel any-angle planner based on Safe Interval Path Planning (SIPP) algorithm is proposed to find trajectories for an agent moving amidst dynamic obstacles (other agents) on a grid. This algorithm is then used as part of a prioritized multi-agent planner AA-SIPP(m). On the theoretical side, we show that AA-SIPP(m) is complete under well-defined conditions. On the experimental side, in simulation tests with up to 250 agents involved, we show that our planner finds much better solutions in terms of cost (up to 20\%) compared to the planners relying on cardinal moves only.

\end{abstract}

\section{Introduction}

Robustness, efficiency and safety of autonomous multi-robot systems used for transportation, delivery, environment monitoring, etc., clearly depends on a) the ability of individual mobile robot to plan its trajectory to the goal b) the ability of robots to avoid collisions. From an AI planning perspective these two tasks can be combined into a single problem of cooperative or multi-agent pathfinding inside a shared environment (workspace), which is typically modeled with a graph. Graph vertices commonly correspond to distinct locations robot(s) can occupy and edges correspond to elementary trajectories, such as line segments, that robot(s) can traverse.
	
Among various graph models, used for both individual and cooperative pathfinding, grids can be named to be the most widespread representations due to their simplicity and seamless integration into the robots' "sense-plan-act" loop.

Typically in 2D grid pathfinding an agent is presumed to move from one traversable (unblocked) cell to one of its eight adjacent neighbors. Sometimes diagonal moves are prohibited, restricting an agent's moves to the four cardinal directions only. Various methods can be utilized to find conflict-free paths for multiple agents that move on a grid in such a way: HCA* \cite{silver2005}, OD+ID \cite{standley2010}, MAPP \cite{wang2011}, M* \cite{WagnerC11}, CBS \cite{sharon2015} and it's modifications \cite{barer2014,cohen2016} etc. Some of these methods have initially been developed for grid worlds (like MAPP), while others (like CBS) can apply to arbitrary graphs (grids including).	

At the same time, the limitations of 8 (or 4) connected grids have led to increased popularity of any-angle pathfinding. In any-angle pathfinding, an agent is allowed to move into arbitrary directions and a valid move is represented by a line segment, whose endpoints are tied to the distinct grid elements (either the center or the corner of the cells) and which does not intersect any blocked cell. Single agent any-angle pathfinding algorithms like Theta* \cite{nash2007}, Anya \cite{harabor2016} etc. tend to find shorter and more realistic looking paths, e.g. paths without numerous heading changes.

Incorporating any-angle planners into the cooperative path-finding framework is a non-trivial task for the following reason. Even in the simplest case when circular agents of the identical size (equivalent to the size of the grid cell) move with identical speeds in 2D environment, following any-angle paths, there is no one-to-one correspondence between the grid elements and possible conflict locations (see Figure 1). One needs to resolve this ambiguity as graph-based multi-agent planners in general rely on the fact that conflicts are tied to the vertices or edges of the given graph or grid.

In addition, when multiple agents follow any-angle paths the conflicts can occur at any point of the continuous timeline, as opposed to the case when agents are confined to four cardinal moves on a grid only and conflicts appear at discrete time points 1, 2, 3, etc., which allows effortless management of the wait actions.

It is also worth noting here that the problem of collision-free paths for multiple circular robots moving amidst polygonal obstacles in general is NP-hard \cite{spirakis1984}.

\begin{figure}[ht]
\centering
\includegraphics[scale=1.2]{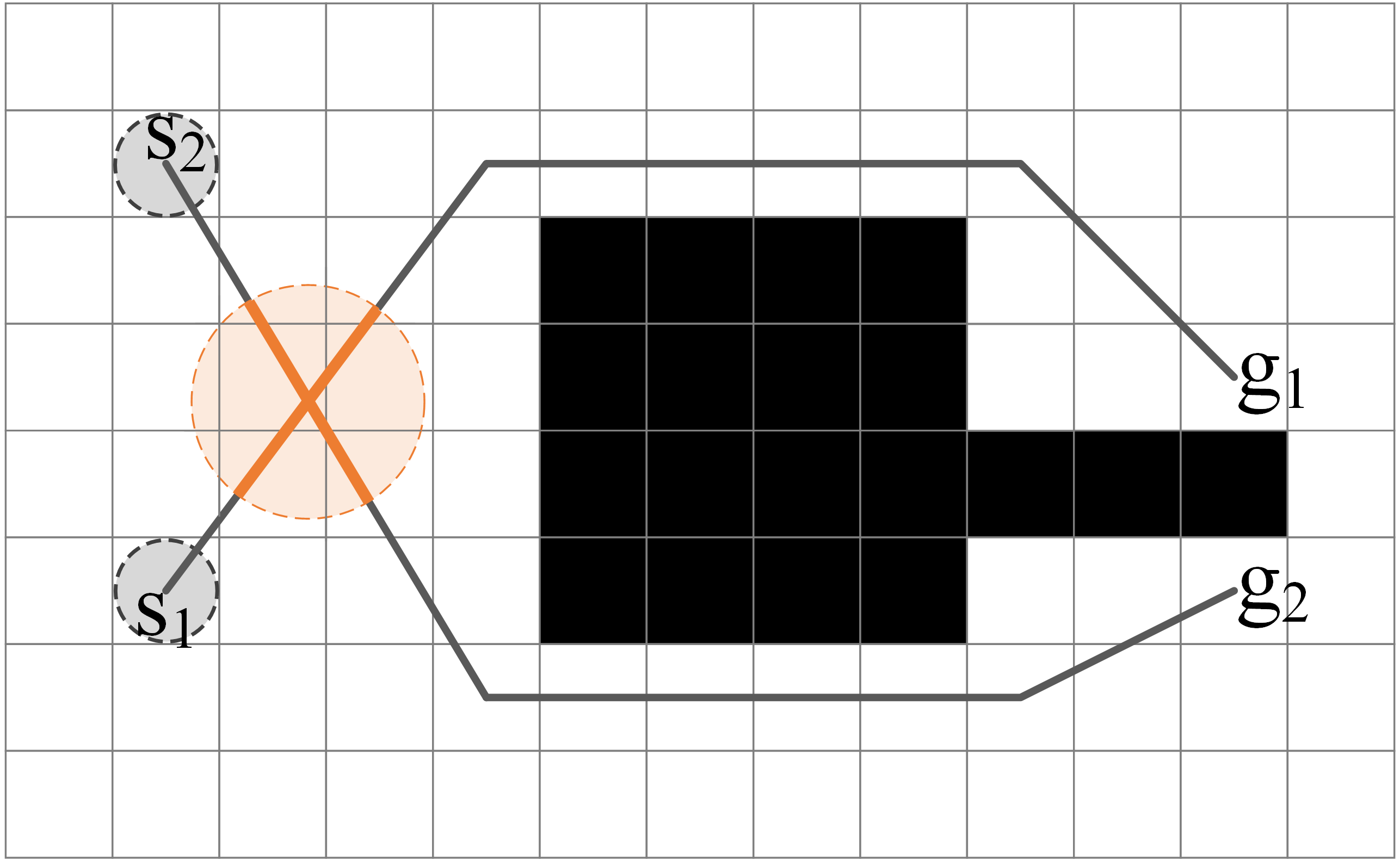}
\caption {Continuous conflict zone (shaded circle) for any-angle paths on grid.}
\label{pic1}
\end{figure}

To attack the aforementioned challenges we suggest a) using decoupled, e.g., prioritized, approach to cooperative pathfinding; as well as b) using any-angle adaptation of the SIPP algorithm \cite{phillips2011} to search for the individual paths.

In prioritized planning \cite{erdmann1987} all agents are assigned unique priorities and paths are planned one by one in accordance with the imposed ordering using some (preferably complete) algorithm. From previous research, it is known that decoupled planners scale well to large problems and are much more efficient computationally-wise than the coupled ones that search for a solution in joint state space which is a cardinal product of individual state spaces. On the other hand, the latter planners are optimal while the former are not even complete in general. At the same time, under certain conditions their completeness can be guaranteed \cite{cap2015a} and these conditions typically hold in practice. Thus, decoupled planning is an appealing alternative to the coupled approach, especially from a practical point of view. Motivated by this, we have developed a new decoupled, prioritized planner, AA-SIPP(m) (any-angle SIPP applied to multiple agents), capable of solving cooperative pathfinding problems under any-angle assumption. To the best of authors' knowledge, it is the first algorithm of this kind.
	
	The proposed method is complete under well-defined conditions, as well as highly efficient in practice. The results of the experimental evaluation show that the success rate of AA-SIPP(m) is extremely high ($>$97\%) and the average solution cost is significantly better (up to 20\%) than the one achieved by both coupled and decoupled planners, that rely on cardinal-only moves.

\section{Problem Statement}

We first formulate the problem of cooperative pathfinding in $\Re^{2}$, following the notation from \cite{cap2015a}, and then convert it to a grid-based search problem. 

Consider $n$ circular agents of identical radii, $r$, and identical maximum moving speeds, $v$, operating in 2D workspace, ${W}\subset{\Re^{2}}$, populated with arbitrary obstacles. The corresponding configuration space is $C = C_{\text{free}} \cup C_{\text{obs}}$, where $C_{\text{free}}$ is the free space and $C_{\text{obs}} = C  \backslash C_{\text{free}}$. The task of each agent {\it i} is to move from it's start location, $s_{i} \in C_{\text{free}}$, to it's goal location, $g_{i} \in C_{\text{free}}$. This task can be accomplished by following the {\it trajectory}, which is a mapping from the time points to the locations in free space, $\pi_{i}: [0, +\infty) \rightarrow C_{\text{free}}$. The trajectory is {\it feasible} if it starts at the agent's start location: $\pi(0)=s_{i}$, reaches and stays at the goal location: $\exists t_{g}: \pi(t)=g_{i} \forall t>t_{g}$, and the agent following $\pi$ never moves faster than its maximum speed. The trajectories $\pi_{i} \text{,} \pi_{j}$ of two agents are said to be conflict-free if the bodies of the agents never intersect when they follow the trajectories. 

\textbf{Problem}. Given the workspace $W$, as well as the configuration space $C$, and the set of start and goal locations $\{s_{1}, \ldots, s_{n}\}, \{g_{1}, \ldots, g_{n}\}$ find a set of feasible trajectories $\{\pi_{1},\ldots, \pi_{n}\}$ such that each pair of them is conflict-free.
	To convert this problem to the graph-search problem we assume the following:
	
	\begin{enumerate}
		\item{$W$ is a bounded rectangle $\{(x, y)$$:$$x_{min}$$\leq$$x$$\leq$$x_{max}$, $y_{min}$$\leq$$y$$\leq$$y_{max}\}$ which is tessellated into the square grid.}
	\item {The size of the grid cell is $2r$. Thus, assuming that an agent is an open disk of radius $r$, the former fits into the cell when the centers of the agent and the cell coincide.}
	\item {Grid cell $c$ is traversable (free) if $c \cap C_{\text{free}}=c$ and is un-traversable (blocked) otherwise.}
	\item {Inertial effects are neglected, e.g. agents start, stop and turn instantaneously and always move with their maximal speed $v$. Without loss of generality, that speed is $2r$ (one cell-width) per unit time.}
	\item {Start and goal locations of the agents are confined to the center of traversable grid cells.}
	\item {An agent may move from the center of one traversable cell to the center of the other following the straight line connecting those cells. Thus, moves into arbitrary directions that result in any-angle spatial paths, are allowed.}
	\item {Agents are able to {\it wait} only at the start points of the segments along their paths and at the goal locations.}
	\end{enumerate}
	
	According to the $\text{5}^{\text{th}}$, $\text{6}^{\text{th}}$ and $\text{7}^{\text{th}}$ assumptions, the trajectory now is the pair $\langle${\it path}, {\it wait-list}$\rangle$, composed of the spatial and time components. Path is a sequence of the line segments (which endpoints are tied to the centers of the grid cells), starting at $s$ and ending with $g$. The {\it feasible path} is such a path that has at least $r$-clearance of the blocked cells, e.g., an agent following such a path will never collide with static obstacles. Wait-list is a list of {\it wait} actions corresponding to the start points of the segments comprising the path (with the last wait action corresponding to the goal location). As before, two trajectories are {\it conflict-free} if the bodies of the agents never intersect when they follow the trajectories.

Due to the fact that distance is equivalent to time and both are measured in the same units (see assumption 4) the cost of the trajectory is the cumulative sum of the lengths of the path segments and the durations of correspondent wait actions. The cost of the {\it joint solution} for $n$ agents, which is a set of pairwise conflict-free trajectories, is the {\it sum of costs} of individual solutions. In this work cost of the solution is not subject to strict constraints (optimal solutions are not targeted), but low cost solutions are preferable.

\section{Approach overview}

The proposed algorithm relies on decoupled, e.g., prioritized, planning. First, each agent is assigned a unique priority. Second, the trajectories for individual agents are planned sequentially from the highest priority agent to the lowest priority one. For each agent, a trajectory is planned such that it avoids both the static obstacles in the environment and the higher-priority agents moving along the trajectories planned in the previous iterations. This is done by the enhanced SIPP algorithm, which was originally presented by \cite{phillips2011}, and is adapted by us for any-angle path finding.

\section{Planning with SIPP}

SIPP algorithm treats trajectories of the high-priority agents as moving obstacles and uses special technique of grouping contiguous, collision-free time points into safe intervals, which are then used to define the duration of the wait actions. We refer the reader to the papers \cite{phillips2011,narayanan2012} for the detailed explanation of the algorithm as well as examples, and now proceed with an overview.

Each state, $s$, of the SIPP state-space is identified by a tuple $s=[cfg, interval]$ with the additional data, such as $[g(s), h(s), parent(s), time(s)]$ also associated with it. Here $cfg$ is the configuration, e.g. agent's particular location (possibly complimented with the additional information on spatial properties such as heading angle etc.), $interval$  is the safe time interval, e.g. the contiguous period of time for a configuration, during which there is no collision and it is in collision one time point prior and one time point after the period. $time(s)$ is the earliest possible moment of time (within the safe interval) an agent can arrive to $cfg$. $g(s)$ is cost of the best path from the start configuration to $cfg$ found so far via the parent node -- $parent(s)$, $h(s)$ is the heuristic estimate of the cost of the path to goal configuration. It is noteworthy that numerous states with the same configuration, but different disjoint safe time intervals, can simultaneously exist in the search-space.
	
SIPP heuristically searches the state-space using A* strategy, e.g. it iterates through the states choosing the one with the lowest $g(s)+h(s)$ value, e.g. $f$-value, and expanding it. Expanding procedure is composed of successors' generation and their $g$-values and $time$ update. $g$-values are updated in a conventional A* fashion. Successors' generation involves iterating through the configurations reachable from the current one (we will refer to such configurations as {\it neighbors}), calculating their safe intervals and estimating the earliest arrival time for each interval. The latter is used for $time$ update. In case $time$ does not fit inside the safe interval the successor is pruned.

The algorithm stops when the search node corresponding to the goal configuration is selected for the expansion. Then the sought conflict-free trajectory can be reconstructed in the following manner. Goal state, $parent(goal)$, $parent(parent(goal))$ etc. are sequentially added to the list until the start state is reached, the list is then reversed. Configurations ($cfg_0, cfg_1, \ldots, cfg_n)$ of the states $(s_0, s_1,\ldots, s_n)$ residing in the resulting list define a spatial component of the trajectory, e.g a path. Consider now two sequential states, $s_i$, $s_{i+1}$, and correspondent earliest arrival times, $time(s_i)$, $time(s_{i+1})$. In case $time(s_{i+1}) > time(s_i) + dur_{move}(cfgi, cfg_{i+1})$, where $dur_{move}$ is the duration of the move, connecting adjacent configurations, a wait action preceding this move is added. It's duration is $dur_{wait}(cfg_i) = time(s_{i+1}) - dur_{move}(cfg_i, cfg_{i+1})$, e.g. after arriving at $cfg_i$ agent stops and waits for the $dur_{wait}$ time points. Thus, both spatial and time components of the sought trajectory are identified.

\setlength\intextsep{0pt}
\begin{algorithm}[h]
\caption{AA-SIPP}
$g(s_{start})=0\text{\; }OPEN=\oslash$\;
insert $s_{start}$ into $OPEN$ with $f(s_{start})=h(s_{start})$\;
\While{$s_{goal}$ is not expanded}
{
	remove $s$ with the smallest $f$-value from $OPEN$\;
	\For{ each $cfg$ in $NEIGHBORS$($s.cfg$)}
	{
		$successors=\text{getSuccessors}(cfg,s)$\;
		\color{blue}{\If{$cfg$ is reachable from $parent(s).cfg$}
		{
			$successors=successors\cup\text{getSuccessors}(cfg,parent(s))$\;
		}
		}
		\color{black}
		\For{each $s'$ in $successors$}
		{
			\If{$s'$ was not visited before}
			{
				$f(s')=g(s')=\infty$\;
			}
			\If{$g(s')>g(s)+c(s,s')$}
			{
				$g(s')=g(s)+c(s,s')$\;
				updateTime($s'$)\;
				insert $s'$ into $OPEN$ with $f(s')=g(s')+h(s')$\;
			}
		}

	}
}
\Fn{\text{getSuccessors}($cfg,s$)}
	{
		$successors=\oslash$\;
		$m\_time=\text{time to reach } cfg \text{ from } s.cfg$\;
		$start\_t=time(s)+m\_time$\;
		$end\_t=endTime(interval(s))+m\_time$\;
		$intervals=$get all safe intervals for $cfg$\;
		\For{each safe interval $i$ in $intervals$}
		{
			\If{$startTime(i)>end\_t$ or $endTime(i)<start\_t$}
			{
			continue\;
			}
			$t$$=$earliest arrival time from $s$ to $cfg$ during interval $i$ with no collisions\;
			\If{$t$ does not exist}
				{continue\;}
			$s'$$=$state of configuration $cfg$ with interval $i$ and time $t$\;
			insert $s'$ into $successors$\;
			
		}
		return $successors$\;
	}
\end{algorithm}

Pseudo-code for SIPP algorithm is shown in Algorithm 1. Lines 7-8 are related to any-angle SIPP, e.g. AA-SIPP, only and are omitted in the original algorithm.

\section{Any-angle SIPP}

Before proceeding with the detailed explanation of the any-angle SIPP algorithm, please note that when inertial effects are neglected (assumption 4) the configuration of the search state is simply the cell (coordinates of the center of the cell (assumptions 5 and 6), to be more precise).

\subsection{Generating successors}

The core routine of the AA-SIPP algorithm, as well as of the original SIPP, is generating successors for a state. AA-SIPP generates successors twice, first in a conventional SIPP manner and then by trying to achieve a particular configuration, adjacent to the state under expansion, from the parent of this state. Thus, a set of AA-SIPP successors for each state contains all SIPP successors plus possibly additional ones. The latter may have smaller $g$-values and arrival times due to the fact that they are reached in a more straightforward way.
	
	To determine whether one configuration is reachable from the other (line 7), a modified Wu's algorithm is used. The latter is a well-known in computer graphics algorithm \cite{wu1991} that identifies pixels (grid cells in our case) that lie along the straight line connecting two endpoints (configurations). We augmented the original algorithm with additional checks to ensure {\it all} cells lying within $r$ distance from the line segment are identified.
	
	To illustrate how both the original and the modified algorithms work consider the example depicted on Figure 2. In this case, the original Wu shifts from left to right and on each step processes one grid column and marks exactly two vertically adjacent cells that are the closest to the line segment connecting the endpoints\footnote{In practice these cells are identified very quickly using only integer calculations.}. By construction these cells have the property that an open-disk agent of radius $r$ will definitely hit them while moving along the line. It is also possible for an agent to hit other cells of the current grid column, so additional checks (not present in the original algorithm) are performed in the following manner. We calculate the distance between the line segment and the bottom-right/upper-left corner of the cell residing on top/bottom of the cell-batch identified earlier\footnote{Additional pruning rules can also be introduced to reduce the number of top/bottom residing cells being checked. We omit the description of these rules for the sake of space but, indeed, our implementation of the algorithm utilizes them.}. If the distance is less than $r$, the corresponding cell is also marked. The algorithm stops when the grid column containing the endpoint of the line segment is processed with all the cells lying within $r$ distance from the line segment being identified. If all these cells are traversable, two configurations are considered to be reachable one from the other, e.g. open-disk agent of radius $r$ can straightly move between them without colliding with static obstacles.
	
	The complexity of the algorithm is $O(n)$, where $n$ is $max(\Delta x, \Delta y)$ and $\Delta x (\Delta y)$ -- is the absolute difference in $x(y)$-coordinates of the endpoints.

\begin{figure}[ht]
\centering
\includegraphics[scale=0.6]{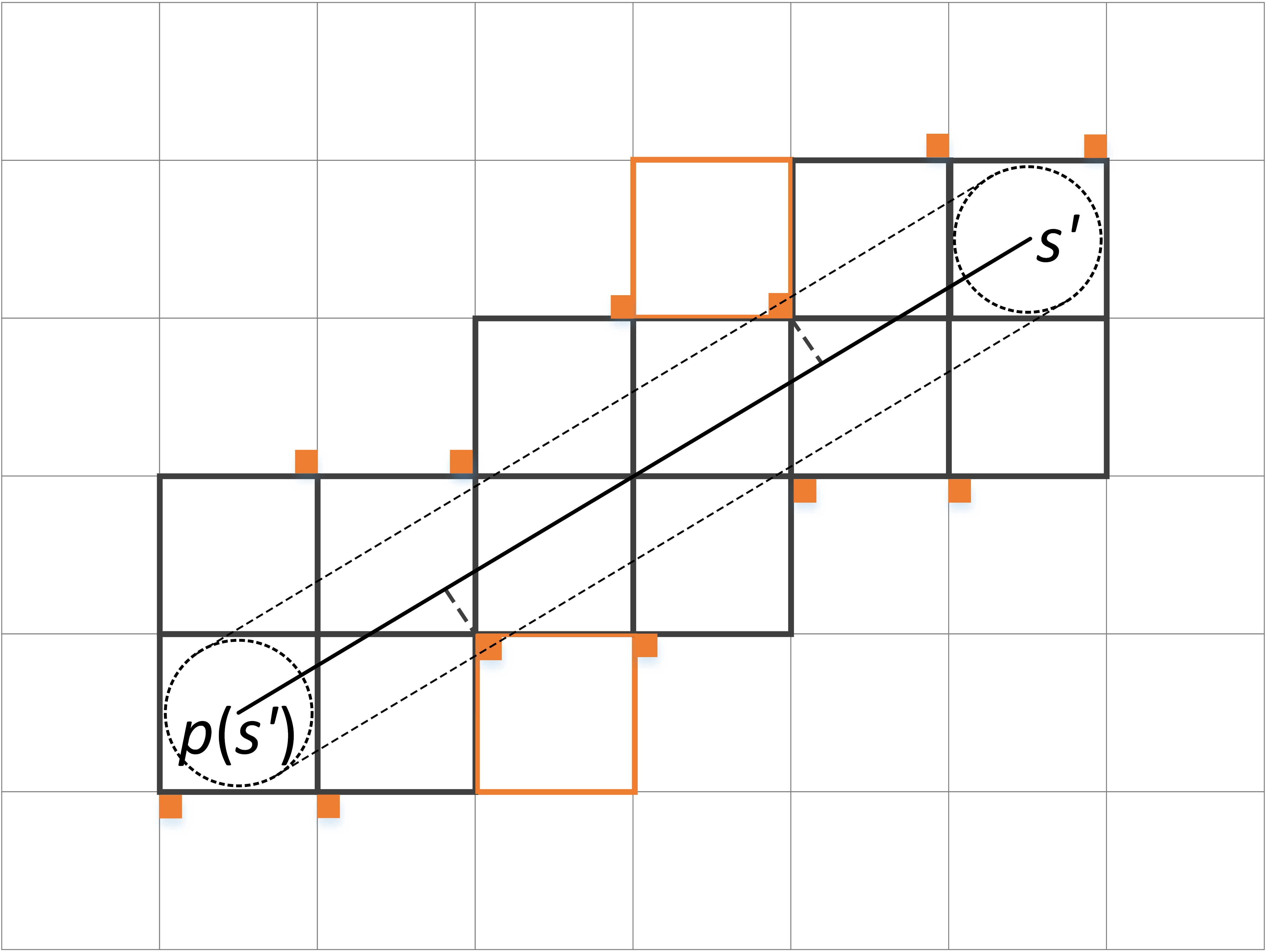}
\caption {Estimating the reachability of the configuration $s'$ from $p(s')$ (best viewed in color). Cells that are identified by Wu's algorithm are highlighted in bold. The ones that are additionally checked are marked with orange square signs in the upper-left/lower-right corners. Some of them (marked with orange border) are additionally added to the result set.}
\label{pic2}
\end{figure}

We now move on to the detailed explanation of how the safe intervals are calculated (line 22) and the earliest available times are estimated (line 25). 

\subsection{Calculation of the safe intervals for a configuration}

Consider a configuration, $cfg$, e.g. center of the grid cell, for which the safe intervals are to be calculated and further used as the part of the search node identifier. We iterate through the paths of the moving obstacles and by invoking modified Wu's algorithm (see above) discover the cells, located within $r$ distance from the paths. If $cfg$ is not among these cells, its interval is $[0, +\infty)$, e.g. not a single moving obstacle hits the cell at any time. In case $cfg$ belongs to the formed set, exact time points defining the start point and the endpoint of the interval are calculated as follows. First, by applying conventional formulas of computational geometry, the coordinates of the intersections of the circumference of radius $2r$ with the center in the interested configuration and the segments of the paths of closely passing obstacles are identified (see Figure 3). Then the arrival times to these points are calculated. These times form the collision interval for the configuration and safe interval is its inversion. In case several moving obstacles pass through the $cfg$, all the safe intervals are calculated and some of them are merged if needed. The latter can happen when several obstacles follow each other with no gap.

\begin{figure}[ht]
\centering
\includegraphics[scale=2]{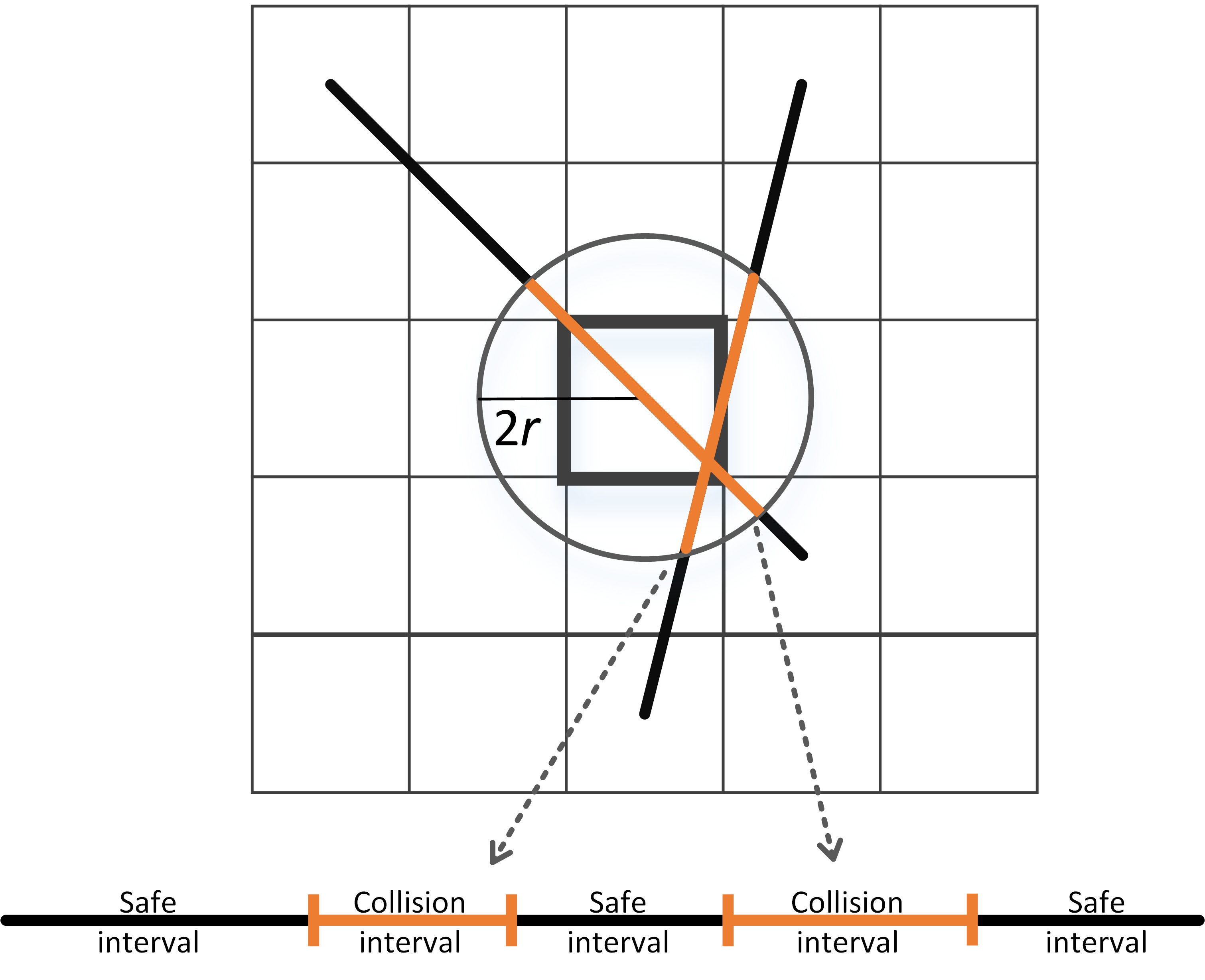}
\caption {Safe and collision intervals for a configuration.}
\label{pic3}
\end{figure}

It's worth pointing out that calculated safe intervals are used only to identify nodes in the search space, not to estimate earliest arrival time or resolve conflicts.

\subsection{Estimating earliest arrival times}

Consider a successor $s'$ of the state $s$ under expansion and a segment $\langle$$parent(s'), s'$$\rangle$. Precise estimation of the earliest arrival time to $s'$ (and how it fits into the current safe interval of $s'$) involves numerous checks and calculations associated with particular formulas of computational geometry and is likely to be very time consuming in practice. Instead we suggest a procedure for the approximate $time(s')$ calculation that takes into account all the time-space constraints imposed by moving obstacles.
	
	We first trace all the segments of obstacles' paths and identify the cells that lie within $r$ distance from them\footnote{On the implementation side this is done only once for each path as soon as it is planned.} (using modified Wu's algorithm as described before) -- $CFG_1$(see Figure 4).  For each identified $cfg \in CFG_1$ we store the coordinates of point $p$, the closest point to the center of the cell $cfg$, lying on the corresponding obstacle path's segment $\langle$$a, b$$\rangle$; as well as $time(p)$, which equals $time(a) + dist(a, p)$. We call this tuple $[p, time(p)]$ -- a constraint. In case $p$ is the start point of $\langle$$a, b$$\rangle$ and obstacle waits in $a$, occupying the cell for a particular time $t_{wait}$ before moving on, multiple constraints $[p, time_1(p)], \ldots, [p, time_k(p)]$ are associated with that cell. Here $time_1(p)=time(a)$, $time_{i+1}=time_i+2r$, $time_k=min\{time_{k-1}+2r, time(a)+t_{wait}\}$.
	
We then invoke modified Wu's algorithm on the end-points of the segment $\langle$${parent(s'), s'}$$\rangle$ and plot the cells lying within $r$ distance from it -- $CFG_2$. We are now interested only in those configurations that are present in both sets $CFG_1$, $CFG_2$ (these cells can be identified extremely quickly in practice by overlapping two sets). More precisely -- we are interested in the constraints $[p,time(p)]$ which are tied to these configurations. It is noteworthy here that the distance between the consecutive constraint points $p$ does not exceed $2r$.

We discard now the constraints $[p, time(p)]$ having the property $dist(p,$$\langle$$parent(s'),s'$$\rangle$$)$$\geq$$2r$ and end up with so-called relevant constraints. Each of them is a marker saying that some obstacle passes nearby the segment $\langle$$parent(s'), s'$$\rangle$ at particular moment of time thus a potential conflict exists and should be resolved.

\begin{figure}[ht]
\includegraphics[scale=0.85]{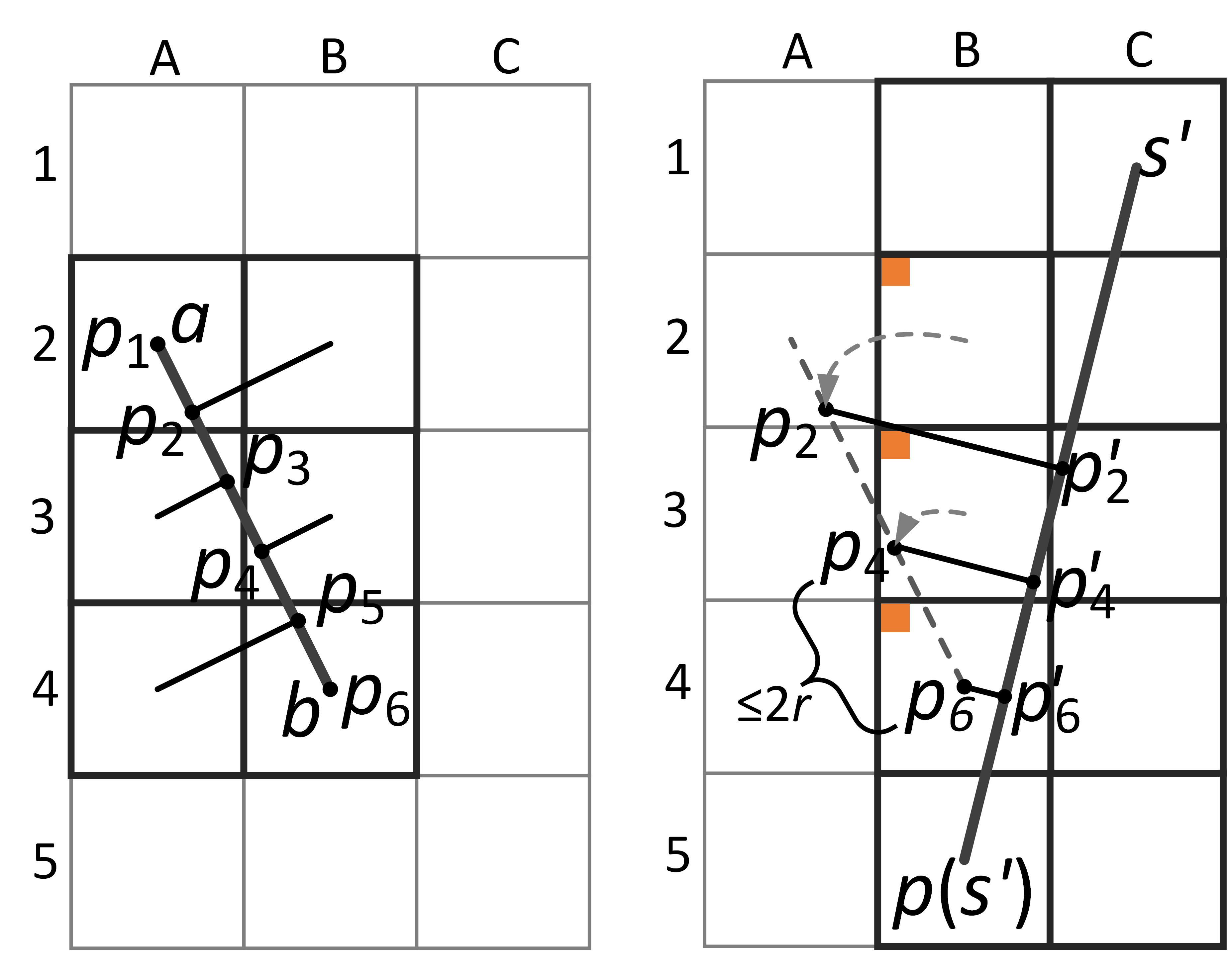}
\caption {Estimating constraints. Left: cells that are hit by the obstacle moving along $\langle$$a,b$$\rangle$ segment are highlighted in bold; $p_1,..., p_6$ -- are the points on the obstacle's path that are the closest to the centers of cells being hit. Right: cells that are hit by the agent are highlighted in bold; cells that are hit by both the obstacle and the agent (B2, B3, B4) are marked with solid squares in the upper-left corners; $p_2', p_4', p_6'$ -- are the points on the agent's path that are the closest to the obstacle path.}
\label{pic4}
\end{figure}

For each constraint $[p, time(p)]$ consider the interval $[time(p) - 4r ; time(p) + 4r]$. This interval is the {\it consistent estimate} of the collision interval at $time(p)$, i.e. an agent moving along any trajectory on a grid is guaranteed not to collide with the obstacle in the $2r$ proximity of point $p$ in case it's trajectory always has $4r$ gap to $[p,time(p)]$. By saying ``having the $4r$ gap to $[p,time(p)]$" we mean that an agent passes the closest to $p$ point on its path either $4r$ time units before or after $time(p)$. The corresponding collision interval for $s'$ will now be $[time(p) - 4r + offset ; time(p) + 4r + offset]$, $offset = dist(s', p')$, where $p'$ is closest to $p$ point on the segment $\langle$$parent(s'), s'$$\rangle$. If an agent starts moving from $parent(s')$ towards $s'$ at any time point outside the collision interval, it is guaranteed not to collide with the obstacle on the segment stretching from $p_{-}$ to $p^{-}$, where $p_{-}$ is the obstacle's position $2r$ time units before $time(p)$ and $p^{-}$ -- is the obstacle's position $2r$ time units after. Given that the distance between any consecutive points in $\{p\}$ does not exceed $2r$, one can claim that using them to calculate the abovementioned intervals result in a consistent overlapping set of collisions for $s'$.

	When all collision intervals (induced by all relevant constraints) for $s'$ are calculated they are used to estimate the earliest arrival time within the interval $i$ (line 25 of the pseudocode) in the following way. First, in case some collision intervals overlap, they are merged. Then we iterate through the resulting set of the collision intervals, $cols$, and check whether $start\_t$ belongs to $cols_i$. If yes, then the sought earliest arrival time, $time(s')$, equals the end of $cols_i$ and we break the cycle. After breaking, we check whether $time(s') > endTime(i)$. If yes, then interval $i$ is unreachable and the successor $s'$ with this interval is pruned. If no, then successor $s'$, defined by the configuration $s'.cfg$ and time interval $i$, with the earliest arrival time $time(s')$ is valid and is inserted into the successors set (lines 28-29).
	
	It also can happen that $start\_t$ doesn't belong to any collision interval from $cols$. It, therefore, means that $time(s')=start\_t$, i.e. an agent can immediately start moving towards $s'$ from $parent(s')$ without waiting.   

\section{AA-SIPP and AA-SIPP(m) properties}

AA-SIPP stands for the any-angle SIPP, and when it is applied as the individual planner inside a prioritized multi-agent planning framework, we end up with algorithm called AA-SIPP(m). These algorithms have the following properties:

\textbf{Property 1}. AA-SIPP is complete (with respect to the grid discretization of the 2D workspace).

\textit{Sketch of proof}. AA-SIPP generates exactly the same successors SIPP does plus possibly some extra ones (shortcuts) and has the same strategy of state space exploration as SIPP. Taking into account SIPP's completeness \cite{phillips2011} we can directly infer the completeness of AA-SIPP.

\textbf{Property 2}. The cost of the AA-SIPP solution never exceeds the cost of the SIPP solution.

\textit{Sketch of Proof}. The claim flows out of the fact that successors set for AA-SIPP is the superset for SIPP successors.

\textbf{Property 3}. AA-SIPP(m) is complete (under assumptions made) in well-defined infrastructures where the well-defined infrastructure is a multi-agent trajectory planning instance having the following property: For any start and goal locations (centers of grid cells), a path exists between them:
  
	a) with at least $r$-clearance with respect to the static obstacles;
	
	b) with at least $2r$-clearance to any other start or goal location.
	
	Informally, a well-formed infrastructure (WFI) has its endpoints distributed in such a way that any agent standing on an endpoint cannot completely prevent other agents from moving between any other two endpoints.
	
	\textit{Sketch of Proof}. It has been shown before that prioritized planner relying on the complete individual pathfinding algorithm that takes into account previously found trajectories as fixed obstacles in space-time, is complete in WFI \cite{cap2015a,cap2015b}. Thus, as AA-SIPP is complete, AA-SIPP(m) is also complete.

\section{Experimental analysis}

We compared the proposed multi-agent planner AA-SIPP(m) with the following grid-based planners relying on cardinal moves only: SIPP(m), ICBS and ECBS. The latter two are the coupled planners of the CBS family: ICBS is the enhanced version of CBS that guarantees finding optimal solutions for cooperative pathfinding problems  \cite{boyarski2015b}; ECBS is the modification of CBS which trades off optimality for speed \cite{barer2014}. We used the implementation of ICBS from the authors' repository\footnote{https://bitbucket.org/eli.boyarski/mapf}, the implementation of ECBS provided by Pavel Surynek\footnote{http://ktiml.mff.cuni.cz/\texttildelow surynek/research/ecbs/} and our own implementations of SIPP(m), AA-SIPP(m)\footnote{https://github.com/PathPlanning/AA-SIPP-m}. The experiments were conducted on Windows-8.1 PC with AMD FX-8350(4.0 GHz) CPU and 16 Gb of RAM. Time limit for each run was set to 5 minutes, e.g. if the planner was not able to produce a solution within this time it was interrupted and the instance was counted as {\it failed} when calculating success rate.

It is worth noting here that all algorithms involved in the experimental analysis are somewhat parameter specific, e.g. they can be parameterized in different ways and different parametrizations might lead to different results. The key parameter for the prioritized planners, e.g. SIPP(m) and AA-SIPP(m), is the way the priorities are assigned. We used the simplest FIFO scheme, e.g. the agent which appeared to be the first in the input task description was assigned the highest priority. Applying different and more complicated prioritization rules, like random restarts \cite{cohen2016} etc., is an appealing direction for future research. As for ICBS and ECBS, we parametrized them in a way the authors of the algorithms had suggested in their original papers: \cite{boyarski2015b} and \cite{barer2014} respectively.

First, we compared the algorithms on 64x64 grids without any obstacles. Instances with the number of agents equal to 50, 100, 150, 200 and 250 were generated (100 instances per each number of agents). For each instance start and goal locations were chosen randomly, in such a way that the distance between any two of them was not less than $4r$. Thus, each instance was a well-formed infrastructure (see above) and was definitely solvable by prioritized planners, e.g. SIPP(m) and AA-SIPP(m).

\begin{table}
	\centering
		\begin{tabular}{|@{}c@{}|@{}c@{}|@{}c@{}|@{}c@{}|@{}c@{}|@{}c@{}|}
		\hline
		\phantom{\'I}\small{N}\phantom{\'I}	& &	\small{ICBS} & \small{SIPP(m)}	& \small{\phantom{.}AA-SIPP(m)\phantom{.}}& \small{ECBS} \\
		\hline
		\multirow{4}{*}{50}  &\phantom{\'I}Success\phantom{\'I} &97.00\%&100.00\%& 100.00\% & 100\%\\
		& Time(s)&3.4176&0.0155& 0.1311&0.1421\\
		&\multirow{2}{*}{Cost}&\multirow{2}{*}{\phantom{.}2240.56\phantom{.}}&2249.31&1758.29 & 2241.76\\
		&&&\small{\phantom{.}(+0.39\%)\phantom{.}}& \small{(-21.52\%)}&\small{\phantom{.}(+0.05\%)\phantom{.}}\\
		\hline
		\multirow{4}{*}{100} &\phantom{\'I}Success\phantom{\'I}& 62.00\%&100.00\%& 100.00\%&100\%\\
		&Time(s)&26.9789&0.0553& 0.4186&0.5905\\
		&\multirow{2}{*}{Cost}&\multirow{2}{*}{4446.65}&4483.66&3575.87&4452.28\\
		&&&\small{(+0.83\%)}& \small{(-19.58\%)}&\small{(+0.13\%)}\\
		\hline
		\multirow{3}{*}{150} &\phantom{\'I}Success\phantom{\'I}&14.00\%&100.00\%& 100.00\%&100\%\\
		&Time(s)&--&0.1236& 0.8931&1.6989\\
		 &Cost&--&6749.81&5485.56&6668.04\\
		\hline
		\multirow{3}{*}{200} &\phantom{\'I}Success\phantom{\'I}&0.00\%&100.00\%& 100.00\%&100\%\\
		&Time(s)&--&0.2147& 1.6784&4.1229\\
		&Cost&--&9106.2&7574.8&8947.8\\
		\hline
		\multirow{3}{*}{\phantom{.}250\phantom{.}} &\phantom{\'I}Success\phantom{\'I}&0.00\%&100.00\%& 100.00\% &100\%\\
		&Time(s)&--&0.3344& 2.8774&9.0114\\
		&Cost&--&\phantom{.}11532.85\phantom{.}&9812.7&10778.6\\
		\hline
		\end{tabular}
	\caption{Results of the algorithms' evaluation on 64x64 empty grids. }
	\label{tab1}
\end{table}

\begin{table*}
	\centering
		\resizebox{\textwidth}{!}{%
		\begin{tabular}{|@{}c@{}|@{}c@{}|@{ }c@{ }|@{ }c@{ }|@{ }c@{ }|@{ }c@{ }|@{ }c@{ }|@{}c@{}|@{}c@{}|@{}c@{}|@{}c@{}|@{}c@{}|@{}c@{}|@{}c@{}|}
		\hline
		\multicolumn{2}{|c|}{ } & \multicolumn{4}{|c|}{\phantom{\'I}brc202d\phantom{\'I}} & \multicolumn{4}{|c|}{den520d} & \multicolumn{4}{|c|}{ost003d} \\
		\hline
		N & &\small{\phantom{\'I}ICBS\phantom{\'I}}&\small{SIPP(m)} & \small{AA-SIPP(m)}& \small{ECBS}& \small{ICBS} & SIPP(m) & \small{\phantom{.}AA-SIPP(m)\phantom{.}}& \small{ECBS} & \small{ICBS} & \small{SIPP(m)}& \small{\phantom{.}AA-SIPP(m)\phantom{.}} & \small{\phantom{.}ECBS\phantom{.}}\\
		\hline
		\multirow{4}{*}{ 25 }  	&\phantom{\'I}Success\phantom{\'I}&100\%&100\%&100\%&100\%&99\%&99\%&99\%&99\%&99\%&100\%&100\%&100\% \\
														&Time(s)												&0.1105&0.0881&0.4735&0.2128&0.347&0.108&0.6873&0.2895&0.6365&0.0966&0.4814&0.1809 \\
														&\multirow{2}{*}{Cost}					&\multirow{2}{*}{ 3242.71}&3244.13&2818.07&3243.12&\multirow{2}{*}{ 3484.85}&3487.18&2808.66&3485.8&\multirow{2}{*}{ 2706.91}&2711.43&2140.75&2708.12\\
														&																&&\phantom{a}(+0.04\%)\phantom{a}&(-13.64\%)&\phantom{a}(+0.01\%)&&\phantom{a}(+0.07\%)\phantom{a}&(-19.13\%)&\phantom{a}(+0.03\%)\phantom{a}&&\phantom{a}(+0.17\%)\phantom{a}&(-20.92\%)&\phantom{a}(+0.04\%)\phantom{a}\\
		\hline
		\multirow{4}{*}{ 50 }  	&\phantom{\'I}Success\phantom{\'I}&98\%&100\%&100\%&100\%&92\%&100\%&100\%&100\%&71\%&99\%&99\%&100\% \\
														&Time(s)												&2.056&0.1929&1.1353&0.4796&4.9273&0.2304&1.7087&0.8292&10.7776&0.2218&1.3025&0.5067\\
														&\multirow{2}{*}{Cost}					&\multirow{2}{*}{ 6389.7}&6394.75&5575.79&6390.99&\multirow{2}{*}{6875.8}&6884.33&5558.98&6877.93&\multirow{2}{*}{5362.21}&5380.37&4273.8&5367.3\\
														&																&&(+0.08\%)&(-13.5\%)&(+0.02\%)&&(+0.12\%)&(-18.91\%)&(+0.03\%)&&(+0.34\%)&(-20.3\%)&(+0.09\%)\\
														
		\hline
		\multirow{4}{*}{ 75 }  	&\phantom{\'I}Success\phantom{\'I}&92\%&100\%&100\%&100\%&75\%&99\%&99\%&100\%&22\%&99\%&99\%&100\% \\
														&Time(s)												&5.2857&0.3128&2.0074&0.9185&15.0511&0.3865&3.2271&1.5346&--&0.4215&2.7905&1.3954\\
														&\multirow{2}{*}{Cost}					&\multirow{2}{*}{ 9600.36}&9611.16&8372.0&9603.35&\multirow{2}{*}{10293.1}&10311.4&8344.37&10298.2&\multirow{2}{*}{--}&\multirow{2}{*}{8373.43}&\multirow{2}{*}{6663.58}&\multirow{2}{*}{8340.22}\\
														&																&&(+0.11\%)&(-13.38\%)&(+0.03\%)&&(+0.18\%)&(-18.66\%)&(+0.05\%)&&&&\\		
		\hline
		\multirow{4}{*}{ 100 }  	&\phantom{\'I}Success\phantom{\'I}&66\%&97\%&97\%&99\%&45\%&100\%&100\%&100\%&4\%&99\%&99\%&100\% \\
														&Time(s)												&13.1123&0.4719&3.2037&1.6128&32.6364&0.5654&4.9068&1.9032&--&0.5931&4.2395&2.5662\\
														&\multirow{2}{*}{Cost}					&\multirow{2}{*}{12806.2}&12822.7&11094.1&12811.2&\multirow{2}{*}{13665.3}&13697.5&11105.4&13674.2&\multirow{2}{*}{--}&\multirow{2}{*}{10959.5}&\multirow{2}{*}{8755.2}&\multirow{2}{*}{10906.4}\\
														&																&&(+0.13\%)&(-13.37\%)&(+0.04\%)&&(+0.24\%)&(-18.73\%)&(+0.07\%)&&&&\\		
		\hline
		\end{tabular}
		}
	\caption{Results of the algorithms' evaluation on grid-worlds from Dragon Age: Origins.}
	\label{tab2}
\end{table*}
	
Averaged results of this test are presented in table 1. When calculating average runtimes and solution costs for 50- and 100-agents only those instances that were successfully handled by all algorithms were taken into account. Unfortunately, in cases where the number of agents exceeds 100, the success rate of ICBS is extremely low (up to 0\%). Thus, the latter was excluded from the comparison and runtime and cost for SIPP(m), AA-SIPP(m) and ECBS were calculated by averaging over all instances (as all of them were solved by these planners under the 5 minutes time cap).

As one can see prioritized planners are faster than the coupled ones, although the provided times should not be compared directly due to the different algorithms' implementations.	Surprisingly there is not that much difference between ICBS, SIPP(m) and ECBS in terms of solution quality, e.g. cost. At the same time, costs of the AA-SIPP(m) solutions are nearly 20\% lower on average than the costs of the cardinal-moves-optimal solutions of ICBS.

Next, we compared algorithms on three grid-worlds from the “Dragon Age: Origins” collection of the well-known Nathan Sturtevant's benchmark set \cite{sturtevant2012}. These were the maps that had previously been used in the community (see the IJCAI-2015 paper on ICBS \cite{boyarski2015b}): brc202d, den520d, ost003d. brc202d has almost no open spaces and many bottlenecks, den520d has many large open spaces and no bottlenecks and ost003d has a few open spaces and a few bottlenecks. Start locations were chosen randomly and respective goal locations were chosen by a 100000-step random walk from the starting ones. There were generated 100 problem instances for 25, 50, 75 and 100 agents. The results of the experiment are presented in table 2.
	
	As before, one can notice that the cost difference between optimal solutions gained by ICBS and solutions provided by ECBS and SIPP(m) is negligible -- less than a percent on average. While the latter planners are notably faster (except in cases where the number of agents is relatively small, e.g. 25). AA-SIPP(m), as before, is faster than ICBS, slower than SIPP(m) and the costs of its solutions are significantly smaller than the costs of optimal-for-cardinal-moves-only ones. For den520d and ost003d cost reduction is 19-20\% on average, while for brc202d it's a bit lower, e.g. 13.5\%. This is perfectly explainable as brc202d map models a corridor-like environment without large open spaces, present on den520d and ost003d, where the paths can be smoothed by any-angle moves.
	
	It's also notable that despite the fact ICBS is guaranteed to find an optimal solution for any multi-agent pathfinding problem, it actually fails to do so very often when the reasonable time limit (5 minutes) is set. While proposed prioritized, any-angle planner, AA-SIPP(m), being generally incomplete, solves almost all problems under imposed time constraints. Thus, one can claim that, practically, mixed planning strategies and practices should be used to achieve best results and that any-angle multi-agent pathfinding should be one of such practices.

\section{Summary}

In this work we have studied a multi-agent pathfinding problem in case square grids are used as the environment model and each agent is allowed to move in arbitrary directions. We adopted the prioritized approach to solve the problem, which is known to be very effective in practice although does not guarantee completeness in general, as opposed to coupled approaches that search in the combined search-state of the multi-agent system. We have proposed a novel any-angle planner for individual pathfinding that builds on top of the well-known SIPP algorithm. Proposed algorithm is complete and the prioritized multi-agent planner utilizing it, AA-SIPP(m), is also complete under well-defined conditions which often hold in practice.
	
	Conducted experiments confirmed that using AA-SIPP(m) leads to solutions with lower costs (up to 20\%) as opposed to those obtained by the optimal coupled planner, relying on cardinal-only moves. Success rate of AA-SIPP(m) under reasonable time constraints exceeds 97\% (at least for the scenarios we were experimenting with), while the same indicator for state-of-the-art coupled planner, ICBS, is significantly lower.
	
	In future, we intend to continue experimenting and modifying AA-SIPP and AA-SIPP(m) to speed up the algorithms. Another appealing direction of research is applying AA-SIPP to 3D workspaces. 
\section{Acknowledgments}
This work was supported by the Russian Science Foundation (Project No. 16-11-00048).
\bibliographystyle{AA-SIPP-ArXiv}
\bibliography{AA-SIPP-ArXiv}

\end{document}